\documentclass[fleqn]{article}
\usepackage{spconf, graphicx, amsmath, amssymb, subfigure, verbatim, url}

\graphicspath{{Images/}}

\emergencystretch=\maxdimen
\newcommand{\argmin}{\mathop{\rm arg~min}\limits}

\title{A Data-Driven Compressive Sensing Framework Tailored for Energy-efficient Wearable Sensing}

\name{Kai XU, Yixing Li, Fengbo Ren}
\address{Arizona State University}

\usepackage[font=small,skip=0pt]{caption}

\begin{document}
\maketitle

\begin{abstract}
Compressive sensing (CS) is a promising technology for realizing energy-efficient wireless sensors for long-term health monitoring. However, conventional model-driven CS frameworks suffer from limited compression ratio and reconstruction quality when dealing with physiological signals due to inaccurate models and the overlook of individual variability. In this paper, we propose a data-driven CS framework that can learn signal characteristics and personalized features from any individual recording of physiologic signals to enhance CS performance with a minimized number of measurements. Such improvements are accomplished by a co-training approach that optimizes the sensing matrix and the dictionary towards improved restricted isometry property and signal sparsity, respectively. Experimental results upon ECG signals show that the proposed method, at a compression ratio of 10x, successfully reduces the isometry constant of the trained sensing matrices by 86$\%$ against random matrices and improves the overall reconstructed signal-to-noise ratio by 15dB over conventional model-driven approaches.
\end{abstract}


\section{INTRODUCTION}
\subsection{Background}
The existing healthcare model based on episodic examination or short-term monitoring for disease diagnosis and treatment suffers from the overlook of individual variabilities and the lack of personal baseline data. Long-term or non-intermittent monitoring is the key to creating the big data of individual health record for studying the variability and obtaining the personal baseline. Recent advances in wireless body area network (WBAN) and bio-sensing techniques have enabled the emergence of miniaturized, non-invasive, cost-effective wireless sensors that can be placed on human bodies for personal health monitoring \cite{guy:systematic}. Through WBAN and Internet, the monitored data can be transmitted to a near-field mobile device for on-site processing, as well as to remote servers for storage and data analysis. These technology advancements will eventually revolutionize the health related services to become more efficient and economical, benefiting billions of individuals.

One of the key challenges faced by the long-term wireless health monitoring is the energy efficiency of sensing and information transfer. Due to the limited battery capacity of wireless sensors, non-intermittent sensing inevitably increases the frequency of battery recharging or replacement, making it less convenient for practical use. In bio-sensing applications, the energy cost of wireless transmission is about two orders of magnitude greater than other components \cite{chen:design}. This implies that reducing the data size for information transfer is the key to improving the energy efficiency of wearable sensors.

Compressive sensing (CS) \cite{candes:compressive} offers a universal and straightforward data encoding scheme that can compress a variety of physiological signals, providing a promising solution to the problem. However, most existing CS frameworks are model-driven and suffer from very limited performance when dealing with physiological signals \cite{Polania:ecg, abo:compression, ren:configurable}. The reasons are two-fold. First, conventional CS frameworks employ random Gaussian or Bernoulli sensing matrices that are generated independently from any data, thereby they fail to leverage any particular geometric structure embedded in the signals of interest. This limits the rank of the sensing matrix required for preserving the Restricted Isometry Property (RIP), leading to limited compression ratio (CR). On the other hand, conventional CS frameworks \cite{polania:compressed, seungjae:mobile, abo:compression} that adopt predetermined basis for reconstruction underestimate the intricacy of philological signals and overlook the criticality of individual variability to signal fidelity, which results in very limited reconstruction performance especially at high CR \cite{ren:configurable}. Our previous study \cite{kai:online} has shown that learned dictionaries can better approximate the underlying statistical model of input data. Therefore, they can significantly improve the sparsity of physiological signals as well as reconstruction performance.

\subsection{Relation to Prior Work}
There have been some recent work on exploiting data structures for compressive sensing \cite{Elad:optimized, Duarte:simultaneous, hegde:numax}. In \cite{Elad:optimized}, the authors aim to minimize the averaged mutual coherence between sensing matrix and dictionary. The major limitation of this work is that the mutual coherence is not a direct indicator of RIP, so the optimization result is not suitable for sensor applications. In \cite{Duarte:simultaneous}, the authors aim to find a sensing matrix $\Phi$ and a dictionary $\Psi$ such that the Gram matrix of the product $\Phi \Psi $ is as close to the identity matrix as possible. The problem is that the Gram matrix can hardly be the identity matrix in practice as $\Psi$ is usually over-complete, so the result is sub-optimal. In \cite{hegde:numax}, the authors aim to preserve the pairwise distance between sample vectors. However, since the NuMax formulation minimizes the transformation distortion against the original signal rather than its sparse coefficient, the trained sensing matrix is not compatible with any over-complete dictionaries. Therefore, these existing approaches are not ideally suitable for the CS of physiological signals in wearable sensing applications.

In this paper, we propose a data-driven CS framework that co-optimizes the sensing matrix and the dictionary towards improved restricted isometry property (RIP) and signal sparsity, respectively, by exploiting the intrinsic data structure of physiological signals. Specifically, online dictionary learning (ODL) \cite{mairal:online} is first adopted to train a personalized basis that further improves signal sparsity by capturing the characteristics and individual variability of physiological signals. Based on the learned dictionary, a distortion minimization problem is formulated to construct a near-isometry and low-rank sensing matrix to guarantee a satisfactory recovery performance at improved compression ratios. Overall, the proposed framework keeps the promise to significantly enhance the reconstruction quality and CR trade-off for the CS of physiological signals.

The data-driven nature of the proposed CS framework is very appealing because it fills the gap between the massive medical data and how to utilize them to improve the quality of sensing. The key insight from this study is that the sensor energy efficiency can be enhanced by learning the intrinsic signal structures from big data through cost-effective computation on server systems, rather than doing costly circuit-level development. Moreover, the proposed data-driven framework is equally applicable to a variety of physiological signals and has the potential to be consistently improved as more and more data is collected for training.

\begin{figure*}
	\centering
	\includegraphics[width=0.75\textwidth]{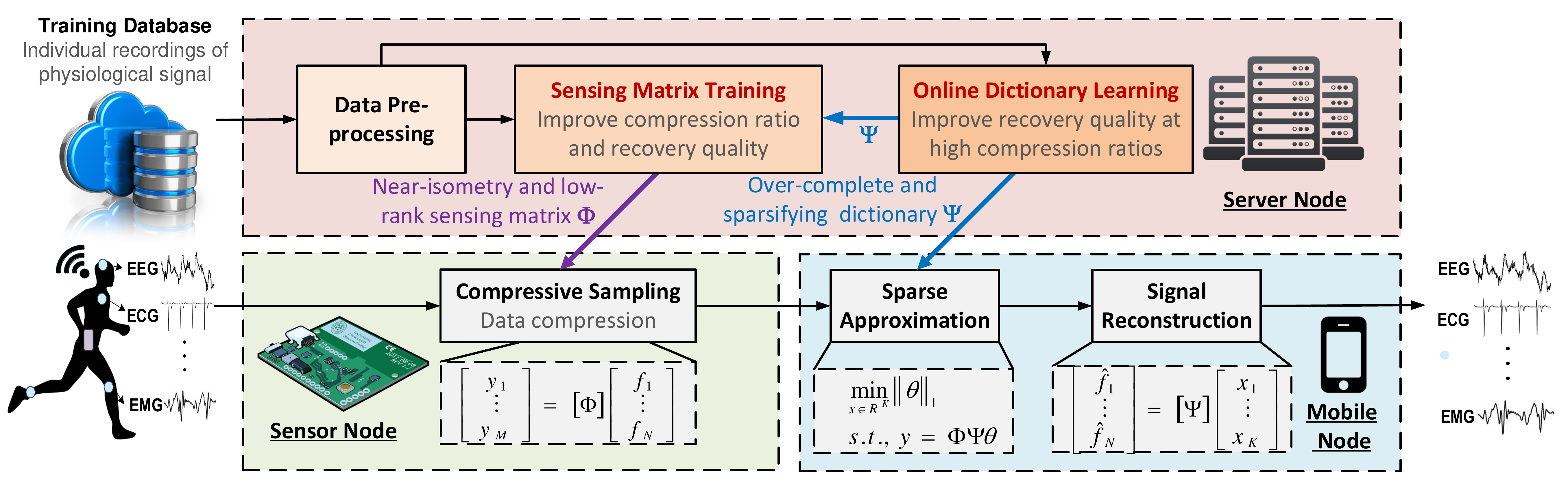}
	\caption{Block diagram of the proposed data-driven compressive sensing framework.}
	\vspace{-0.5cm}
	\label{fig:arch}
\end{figure*}

\section{Preliminaries}
When fully implemented in the digital domain, CS can be considered as a dimensionality reduction technique for signal compression. Assuming a signal $\mathbf{f}$ can be represented by a sparse vector $\mathbf{\theta}\in \mathbb{R}^{k}$ on a certain basis $\Psi \in \mathbb{R}^{n\times k}$, i.e., $\mathbf{f}=\Psi \mathbf{\theta}$, the signal information $\mathbf{x}$ can be well preserved by projecting $\mathbf{f}$ onto a low-dimension space through a sensing matrix $\Phi\in \mathbb{R}^{m\times n}$, ($m\leq n$ and $\Phi$ should satisfy \ref{eq:rip}), given as
\begin{equation}
	\mathbf{y}=\Phi \mathbf{f} + \mathbf{z} =\Phi \Psi \mathbf{\theta} + \mathbf{z},
\end{equation}
where $\mathbf{z}$ is a noise term.

For robust reconstruction, the matrix $\Phi$ should satisfy the RIP \cite{Candes:Incomplete} for all k-sparse signal $\mathbf{x}$, defined as
\begin{equation}
\label{eq:rip}
(1-\delta_{K}\Vert \mathbf{x} \Vert_{2}^{2}\leq \Vert \mathbf{\Phi x} \Vert_{2}^{2}\leq (1+\delta_{K} \Vert \mathbf{x} \Vert_{2}^{2})).
\end{equation}

When the RIP holds, $\Phi$ approximately preserves the Euclidean norm of all k-sparse signals. Then the sparse coefficient can be solved the following $\ell$-1 minimization problem with a relaxed constraint,
\begin{equation}
\label{eq:eq11}
\min_{\mathbf{\theta}\in \mathbb{R}^{k}}{\Vert{\mathbf{\theta}} \Vert_{1}} \quad s.t.\quad \Vert \mathbf{y-\Phi \Psi \theta} \Vert_{2} \leq \varepsilon.
\end{equation}

If matrix $A \in \mathbb{R}^{m \times n}$ satisfies the RIP of order 2k with $\delta_{2k} < \sqrt{2}-1$, the solution to \ref{eq:eq11} is equivalent to the original signal with overwhelming probability \cite{Candès:property}. In addition, we have 
\begin{equation}
\label{eq:eq8}
\| \mathbf{x^{\ast}} - \mathbf{\hat{x}} \|_{\ell 2} \leq C \cdot \frac{\| \mathbf{x - x_{k}} \|_{\ell 1}}{\sqrt{k}},
\end{equation}
where $\mathbf{x} \in \mathbb{R}^{n}$ is the input signal, $\mathbf{x}_{k}$ is the k-sparse approximation, and $\mathbf{\hat{x}}$ is the solution to \ref{eq:eq11}, and C is a constant which is proportional to the isometry constant $\delta_{2k}$.
Eq. \ref{eq:eq8} means a smaller isometry constant guarantees a smaller recovery error, which is suitable for target applications.

\section{FRAMEWORK ARCHITECTURE}
\subsection{Architecture Overview}
The architecture of the proposed framework is shown in Fig. ~\ref{fig:arch}. It is composed of three functional units, including a training unit, a CS sampling unit and signal recovery unit performed on server, sensor and mobile nodes, respectively.

Since physiological signals can vary among different patients, a generic basis for all patients usually perform poorly. The dictionary learning module trains personalized basis that captures individual-specific features that are critical to CS recovery, which guarantees a higher sparsity than predetermined basis. Here we employ ODL as the method for dictionary learning. The most notable advantage of ODL is that it does not rely on the matrix factorization upon the entire training data. As a result, the computational complexity is much less compared to the non-online approaches especially for handling large training data. Before ODL is performed, the raw physiological signals must be pre-processed to remove baseline wandering and high-frequency interference. This is essential to achieving a high signal reconstruction quality. Once the dictionary is learned, it can be downloaded to the mobile node to perform accurate signal recovery.

In the proposed framework, the sensing matrix training (SMT) generates a data-specific sensing matrix with minimized rank and a small isometry constant. A small rank further reduces the data size for transmission, and a smaller isometry enhances reconstruction quality denoted by \ref{eq:eq8}. Once the sensing matrix is trained, it can be downloaded to the sensor node to perform effective compression of physiological signals for energy-efficient sensing and information transfer.

\subsection{Sensing Matrix Training (SMT)}
Cand\`{e}s and Tao prove that if the sensing matrix $\mathbf{\Phi}$ satisfies the RIP, then $\ell$-1 minimization algorithms can successfully recover a sparse signal from noisy measurements \cite{Candes:Incomplete}. Here we formulate an optimization problem that directly optimizes the RIP towards lower isometry constant $\delta$ and lower rank of the sensing matrix $\Phi$ in \ref{eq:eq3}.
\begin{equation}
\label{eq:eq3}
	(1-\delta) \Vert \mathbf{\theta} \Vert_{2} \leq \Vert \mathbf{\Phi \Psi \theta} \Vert_{2} \leq (1+\delta) \Vert \mathbf{\theta} \Vert_{2},
\end{equation}
where $\theta$ is the sparse coefficient vector under the dictionary $\Psi$.
	\ref{eq:eq3} is equivalent to
\begin{equation}
\label{eq:eq2}
	\vert \Vert \mathbf{\Phi \Psi \theta} \Vert_{2} - \Vert \mathbf{\theta} \Vert_{2} \vert \leq \delta,
\end{equation}
when $\mathbf{\theta}$ is normalized.

Suppose we have $L$ sparse coefficients, $\theta_i, i = 1,\ldots, L$ , the optimization problem is essentially to guarantee each of them will satisfy \ref{eq:eq2}, which can be then reformulated as

\begin{equation}
\label{eq:eq1}
\vert \mathbf{\theta}_{i}^{T}(\mathbf{\Psi}^{T} \mathbf{\Phi}^{T} \mathbf{\Phi \Psi - I})\mathbf{\theta}_{i} \vert \leq \delta, \quad i=1,\ldots, L.
\end{equation}

Assume $\mathbf{A = \Phi \Psi}$, $\mathbf{Y}=\mathbf{A}^{T}\mathbf{A}$,  \ref{eq:eq1} can be represented as
\begin{equation}
	\vert \mathbf{\theta}_{i}^{T}(\mathbf{Y - I})\mathbf{\theta}_{i} \vert \leq \delta, \quad i=1,\ldots, L.
\end{equation}

As the rank of the sensing matrix implies the data size for transmission after compression, we also aim to minimize the rank of the sensing matrix in \ref{eq:eq4}. Since the rank minimization problem is not convex, we use the nuclear norm as a proxy to relax the problem to \ref{eq:eq5}.
\begin{equation}
\label{eq:eq4}
\begin{aligned}
	& \underset{\mathbf{Y}}{\text{min}} & &  \{ \vert \mathbf{\theta}_{i}^{T}  (\mathbf{Y-I})  \mathbf{\theta}_{i} \vert, \: \text{rank} (\mathbf{Y}) \},  \quad i = 1,\ldots, L \\
	& \text{s.t.} & & \mathbf{Y} \succeq 0, \\
	& & & \text{diag} (\mathbf{Y})=[1, \; 1 \;, \ldots, \; 1]^{T}.
\end{aligned}
\end{equation}

\begin{equation}
\label{eq:eq5}
\begin{aligned}
	& \underset{\mathbf{Y}}{\text{min}} & &  (\delta +\beta \Vert \mathbf{Y} \Vert^{\ast}) \\
	& \text{s.t.} & & \mathbf{Y} \succeq 0,  \\
	& & & \text{diag} (\mathbf{Y})=[1, \; 1 \;, \ldots, \; 1]^{T} \\
	& &	& |\mathbf{\theta}_{i}^{T}  (\mathbf{Y-I})  \mathbf{\theta}_{i}| \leq \delta, \quad i=1,\ldots, L.
\end{aligned}
\end{equation}
where $\beta$ is the penalty parameter for the nuclear norm.
Then, we perform an Cholesky decomposition to obtain the matrix $\mathbf{A}$, and a singular value decomposition (SVD) to derive the sensing matrix $\Phi$, as defined in \ref{eq:eq9} and \ref{eq:eq10}, respectively.
\begin{equation}
\label{eq:eq9}
	\mathbf{Y} = \mathbf{A}^{T}\mathbf{A} = \mathbf{USU}^{T}, \mathbf{A} = (\mathbf{U} \text{sqrt}(\mathbf{S}))^{T}
\end{equation}
\vspace*{-0.5cm}
\begin{equation}
\label{eq:eq10}
	\mathbf{\Psi} = \mathbf{USV}^{T}, \mathbf{\Psi}^{\dagger} = \mathbf{V S}^{-1} \mathbf{U}^{T}, \mathbf{\Phi} = \mathbf{A \Psi}^{\dagger}
\end{equation}
\vspace*{-0.5cm}
\subsection{Online Dictionary Learning (ODL)}
We seek the dictionary that gives the best representation of every item in the training dataset under the sparsity constraint. The advantage of learning dictionaries from individual recordings of physiological signals is that it provides much better sparse representations than model-driven approaches by exploiting the rich information embedded in the training data. ODL offers faster training speed and fewer storage requirements because of the online processing nature. It is also possible to add new features to the dictionary without stalling the reconstruction using ODL, which offers a mechanic of melioration when a distinctive input is received. Due to the page limit, we would like to refer the readers to \cite{mairal:online} for details of ODL.

\subsection{Co-training of sensing matrix and dictionary (CTSMD)}
We aim to jointly improve signal sparsity and isometry constant through a co-training approach. The proposed CTSMD algorithm is described in Algorithm 1. One should note that the proposed CTSMD algorithm is a non-iterative process. Empirical results show that one round of CTSMD is sufficient to obtain a well-defined results.
\vspace{14mm}
\vspace{2mm}
\hrule
\vspace{1mm}
\noindent {\bf Algorithm 1} Pseudocode for CTSMD
\vspace{1mm}
\hrule
\vspace{2mm}
\noindent {\bf Input:} $\mathbf{x}\in \mathbb{R}^{n}$, $\mathbf{\Psi_{0}}\in \mathbb{R}^{n \times k}, \lambda, \beta$,

\noindent {\bf Output:} $\mathbf{\Phi}, \mathbf{\Psi}$,

\noindent {\bf Online dictionary learning:}
\vspace{-2mm}
\begin{flalign*}
	 & 1) \quad \mathbf{\theta}_{t} = \argmin_{\theta \in \mathbb{R}^{n}} \frac{1}{2} \Vert \mathbf{x}_{t}-\mathbf{\Psi}_{t-1} \mathbf{\theta} \Vert_{2}^{2}+\lambda \Vert \mathbf{\theta} \Vert_{1}, & \\
	 & 2) \quad \mathbf{\Psi}_{t} = \argmin_{\mathbf{\Psi}} \frac{1}{t} \sum_{i=1}^{t} \frac{1}{2} \Vert \mathbf{x}_{i}-\mathbf{\Psi} \mathbf{\theta}_{i} \Vert_{2}^{2} + \lambda \Vert \mathbf{\theta}_{i} \Vert_{1}, &
\end{flalign*}
\noindent {\bf \text{Sensing matrix training:}}
\vspace{-3mm}
\setlength{\mathindent}{1cm}
\begin{flalign*}
	\noindent	&3) \quad \min_{\mathbf{Y}}(\delta  +\beta \Vert \mathbf{Y} \Vert^{\ast}) & \\
	&s.t. \quad   \mathbf{Y} \succeq 0, \nonumber & \\
	&\qquad \text{diag}(\mathbf{Y})=[1 \; 1 \ldots \; 1], \nonumber & \\
	&\qquad |\mathbf{\theta}_{i}^{T}  (\mathbf{Y-I})  \mathbf{\theta}_{i}| \leq \delta, \quad i=1,\ldots, L, \nonumber & \\
	& 4) \mathbf{Y} = \mathbf{A}^{T}\mathbf{A} = \mathbf{USU}^{T}, \mathbf{A} = (\mathbf{U} \text{sqrt}(\mathbf{S}))^{T}, & \\
	& 5) \mathbf{\Psi} = \mathbf{USV}^{T}, \mathbf{\Psi}^{\dagger} = \mathbf{V S}^{-1} \mathbf{U}^{T}, \Phi = \mathbf{A \Psi}^{\dagger}. &
\end{flalign*}
\hrule

\section{EXPERIMENTS}

\subsection{Experimental setup}
Real electrocardiogram (ECG) data from the MIT-BIH arrhythmia ECG database \cite{PhysioNet} is used to benchmark the proposed framework. The customized solver is used for ODL problem and CVX solver \cite{boyd:cvx} is used to solve the SMT problem. Due to the large memory requirement of CVX, our experiments are subjected to limited problem size, which has cost a certain performance degradation across our algorithm. Here we extract 3600 samples, and each sample has a dimension of 128. 3000 and 600 samples are used for training and testing, respectively. The training data is first used with the CTSMD algorithm to construct the sensing matrix and the reconstruction dictionary, which are then used to perform CS measurement and signal reconstruction on the testing data. Three reference approaches are compared in our experiments, i.e. random Gaussian sensing matrix with trained dictionary by ODL, trained sensing matrix by SMT with predetermined discrete cosine and wavelet transform (DCT-DWT) dictionary, and random Gaussian sensing matrix with a predetermined DCT-DWT dictionary.

$CR = n/m$ and reconstructed signal-noise ratio (RSNR) = $\Vert \mathbf{x} \Vert_{2}/\Vert \mathbf{x-x'} \Vert_{2}$  are used as the performance metrics, where $n$ is the dimension of original signal $\mathbf{x}$, $m$ is the number of measurements, and $\mathbf{x'}$ is the reconstructed signal.

\subsection{Experiment Results}
\begin{figure}
	\centering
	\includegraphics[width=0.5\textwidth]{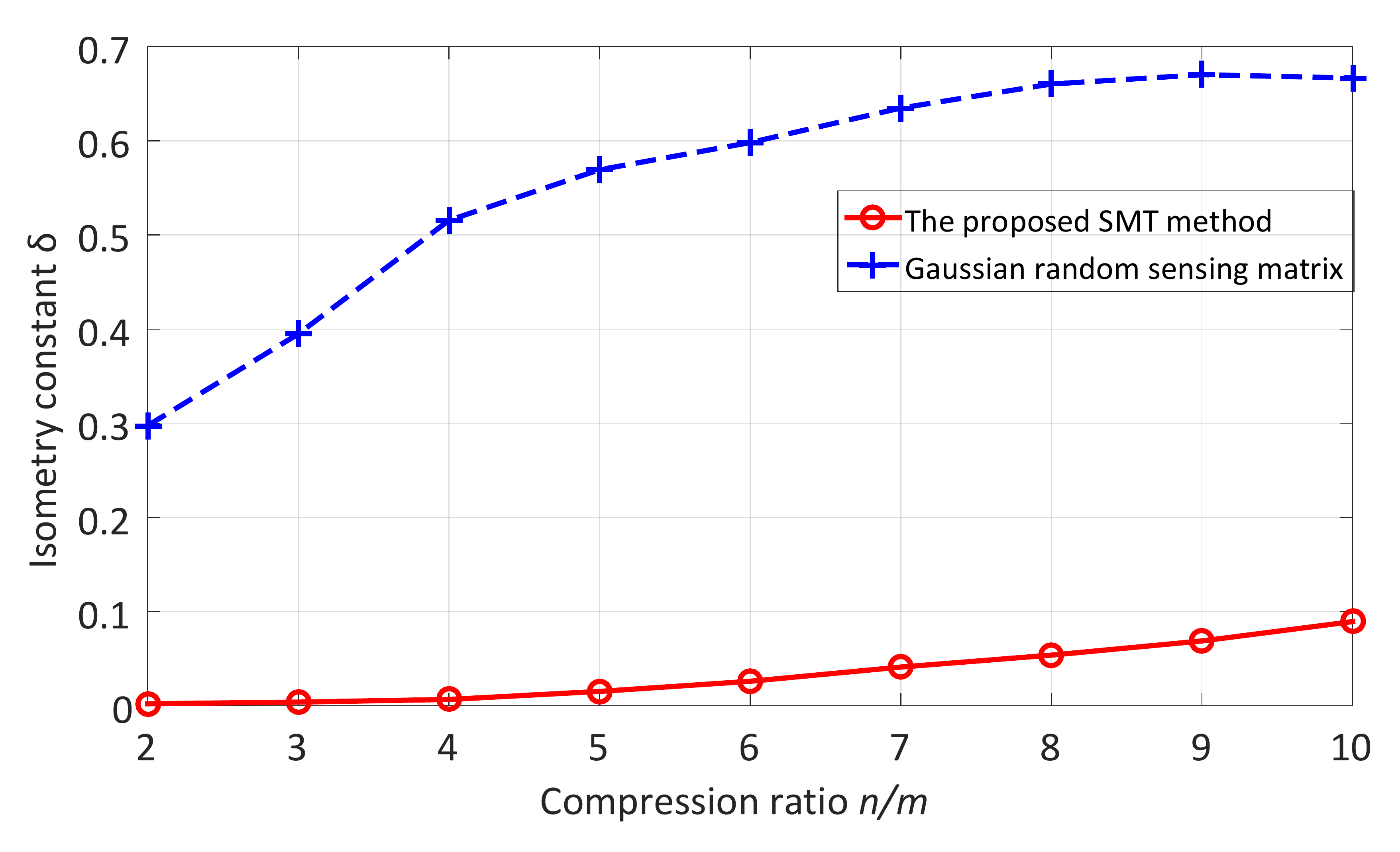}
	\caption{Isometry constant under different compression ratios.}
	\vspace{-5mm}
	\label{fig:is}
\end{figure}

The isometry constant of the trained sensing matrix with respect to CR is shown in Fig. \ref{fig:is}. Note that the sensing matrices produced by the proposed framework have reduced the isometry constant by over 80$\%$ over the Gaussian random matrices across all the CRs. The reduced isometry constant implies better preservation of the signal's geometry structure in the compressed domain. According to \ref{eq:eq8}, such improvement will lead to a higher reconstruction accuracy.

\begin{figure}
	\centering
	\includegraphics[width=0.5\textwidth]{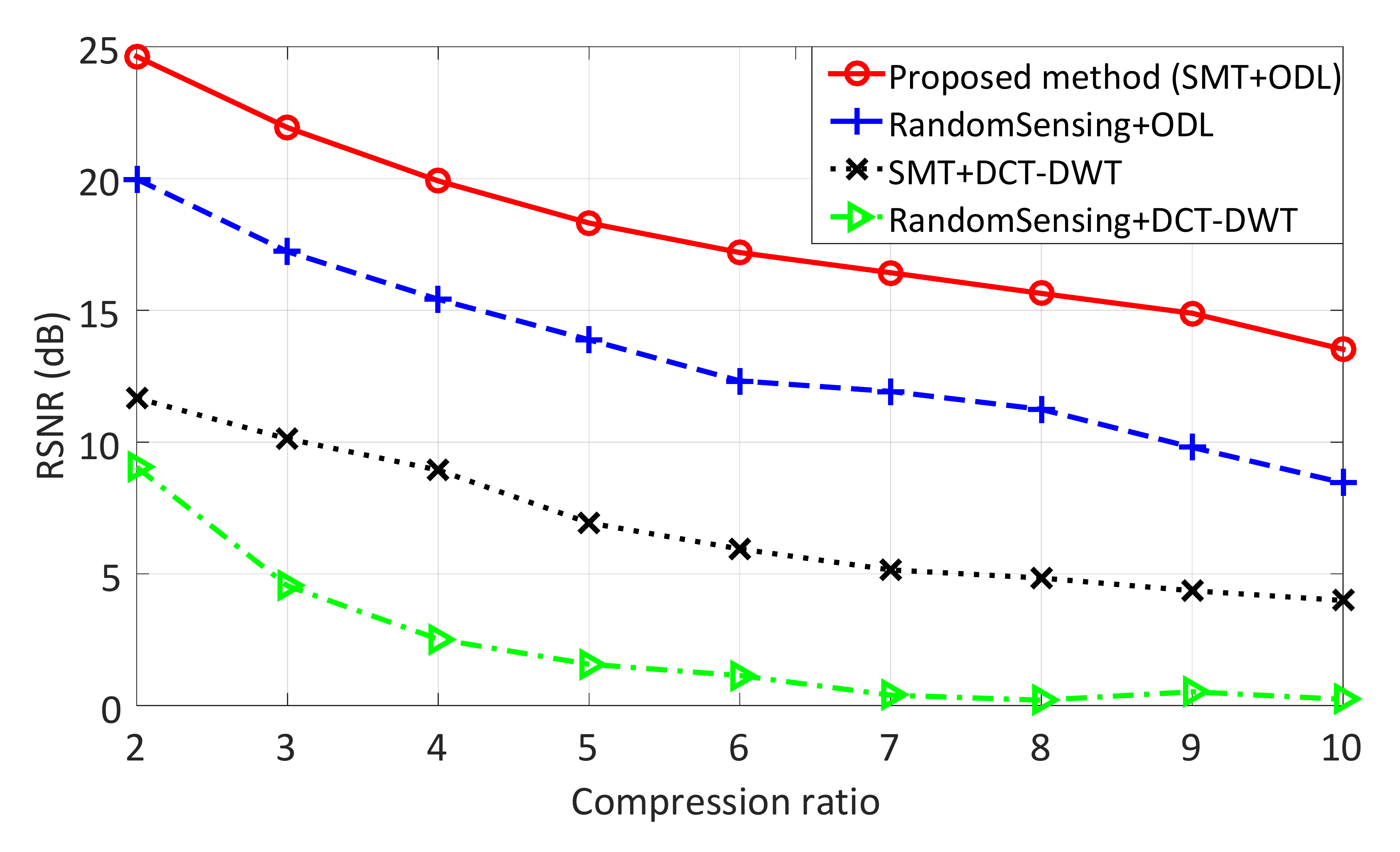}
	\caption{RSNR under different compression ratios.}	
	\vspace{-5mm}	
	\label{fig:rsnr}
\end{figure}

The RSNR results at different CR are shown in Fig. \ref{fig:rsnr}. By using SMT and ODL, RSNR is increased about 5dB and 10dB, respectively. Overall, the proposed data-driven method achieves a 15dB improvement of RSNR over the model-based approach across all different CRs.

\section{CONCLUSION}
In this paper, we propose a data-driven CS framework tailored for the energy-efficient wearable sensing of physiological signals. Exploiting the structure of data is the key to enhancing CS performance. Specifically, the SMT reduces the isometry constant in RIP, and the ODL improves signal sparsity, which are both critical to providing a better recovery performance under improved compression ratios. In future works, we plan to develop customized solver for the SMT problem to handle large dataset. We also need to add binary constraint to SMT for efficient sensor hardware implementations. This will benefit the hardware and energy cost of mobile sensors, which enables the data-driven technique to be used in practical IoTs applications.

\vfill\pagebreak

\bibliographystyle{IEEEbib}
\bibliography{refs}

\end{document}